\documentclass[11pt,a4paper]{article}
\usepackage[hyperref]{emnlp2019}
\usepackage{times}
\usepackage{latexsym}
\usepackage{url}
\usepackage{forest}
\usepackage{qtree}
\usepackage{multirow}
\usepackage{mathtools}
\usepackage{tikz}
\usetikzlibrary{matrix,chains,positioning,decorations.pathreplacing,arrows}
\usepackage{tikz-qtree}
\usepackage{tikz-qtree-compat}
\usepackage[utf8]{inputenc}
\usepackage{xcolor}
\usepackage{pgfplots}
\usepackage{pgfpages}
\usepackage{subcaption}
\usepackage{comment}
\usetikzlibrary{arrows,decorations.pathmorphing,backgrounds,positioning,fit,petri,shapes.misc, arrows.meta,shapes.geometric,decorations.markings,calc,shadows.blur,decorations.pathreplacing,quotes,intersections}
\usepackage{multirow}
\usepackage{array}
\usepackage{arydshln}
\usepackage{mathtools}
\usepackage{makecell}
\graphicspath{ {images/} }
\usepackage{tabularx}
\usepackage{booktabs}
\usepackage{tikz}
\usetikzlibrary{matrix,chains,positioning,decorations.pathreplacing,arrows}
\usepackage{tikz-qtree}
\usepackage{tikz-qtree-compat}
\usepackage[utf8]{inputenc}
\usepackage{xcolor}
\usepackage{pgfplots}
\usepackage{pgfpages}
\usepackage{subcaption}

\usetikzlibrary{arrows,decorations.pathmorphing,backgrounds,positioning,fit,petri,shapes.misc, arrows.meta,shapes.geometric,decorations.markings,calc,shadows.blur,decorations.pathreplacing,quotes,intersections}
\definecolor{fontgrey}{RGB}{44, 62, 80}

\aclfinalcopy % Uncomment this line for the final submission
\newtheorem{definition}{Definition}[section]

\newcounter{Lcount}
\newcommand{\squishenum}{
\begin{list}{\arabic{Lcount}. }
{ \usecounter{Lcount}
\setlength{\itemsep}{0pt}
\setlength{\parsep}{0pt}
\setlength{\topsep}{0pt}
\setlength{\partopsep}{0pt}
\setlength{\leftmargin}{2em}
\setlength{\labelwidth}{1.5em}
\setlength{\labelsep}{0.5em} } }

\newcommand{\squishletter}{
\begin{list}{\alph{Lcount}. }
{ \usecounter{Lcount}
\setlength{\itemsep}{0pt}
\setlength{\parsep}{0pt}
\setlength{\topsep}{0pt}
\setlength{\partopsep}{0pt}
\setlength{\leftmargin}{2em}
\setlength{\labelwidth}{1.5em}
\setlength{\labelsep}{0.5em} } }

\newcommand{\squishlist}{
\begin{list}{$\bullet$}
{ \usecounter{Lcount}
	\setlength{\itemsep}{0pt}
	\setlength{\parsep}{0pt}
	\setlength{\topsep}{0pt}
	\setlength{\partopsep}{0pt}
	\setlength{\leftmargin}{2em}
	\setlength{\labelwidth}{1.5em}
	\setlength{\labelsep}{0.5em} } }

\newcommand\blfootnote[1]{%
	\begingroup
	\renewcommand\thefootnote{}\footnote{#1}%
	\addtocounter{footnote}{-1}%
	\endgroup
}

\newcommand{\squishend}{
\end{list} }

\title{\texttt{Text2Math}: End-to-end Parsing Text into Math Expressions}
\author{%
	Yanyan Zou \and Wei Lu\\
	StatNLP Research Group\\
	Singapore University of Technology and Design \\
	{\tt yanyan\_zou@mymail.sutd.edu.sg, luwei@sutd.edu.sg} \\
}

\date{}

\begin{document}
\maketitle
\begin{abstract}

%OK

We propose \texttt{Text2Math}, a model for semantically parsing text into math expressions.
The model can be used to solve different math related problems including arithmetic word problems \cite{roy2017unit,liang2018meaning} and equation parsing problems  \cite{roy2016equation}.
Unlike previous approaches, we tackle the problem from an end-to-end structured prediction perspective where our algorithm aims to predict the complete math expression at once as a tree structure, where minimal manual efforts are involved in the process.
Empirical results on benchmark datasets demonstrate the efficacy of our approach.\blfootnote{This work has been accepted in 2019 Conference on Empirical Methods in Natural Language Processing and 9th International Joint Conference on Natural Language Processing as a full paper.}
%Automatically solving math word problems requires understanding the meaning of problem text to capture mathematical relations (i.e., addition, subtraction, multiplication, and division) among numbers appearing in the text and unknown variables that refer to phrases of the text.
%In this work, we propose a novel approach, \emph{Text2Math}, that solves the math word problems from a semantic parsing perspective which can generate answers with interpretable explanations.
%We design a semantic parsing algorithm with a joint representation that allows to capture information interactions between text and expression trees and to utilize structure knowledge from expression trees.
%Empirical results on benchmark datasets prove the efficacy of our model.
%We make our code available at *URL*.
\end{abstract}

\section{Introduction}

%OK

%It has been proved that a

%This motivates us to build an end-to-end model that maps from a text sequence to a tree representation. 

%To solve an arithmetic word problem, exemplified by Problem 1 in Figure \ref{fig:examples}, an algorithm needs to understand the description of the text and then model the relations through mathematical operations of quantities, consisting of numbers appearing in the text.
%The goal is to calculate the final solution according to the problem description and question.
%Equation Parsing \cite{roy2016equation} is a general case of math word problems exemplified by Problem 2 in Figure \ref{fig:examples}, where the expression\footnote{In this work, we use the term ``expression" to refer to a mathematical expression (like the one for Problem 1 in Figure \ref{fig:examples}) or an equation (like the one for Problem 2 in Figure \ref{fig:examples}).} contains numerical numbers appearing in the problem description and unknown variables mapped to noun phrases residing in the text.
%The mathematical expressions can be equivalently represented by tree structures.
%such as expression trees, as illustrated in Figure \ref{fig:examples}

\begin{figure}[t]
\centering
\scalebox{0.7}
{\begin{tabular}{p{10.5cm}}
		%			\hline
		\textbf{Problem 1} \\
		%		\hline
		{\em Mike picked 7 apples. Nancy picked 3 apples and Keith picked 6 apples at the farm. In total, how many apples were picked?}\\
		%		\hline 
		\textbf{Expression}\textcolor{white}{ion}   $(7+(3+6))$\\
		
		\scalebox{0.7}{
			\begin{tikzpicture}[node distance=2.0mm and 2.0mm, >=Stealth, 
			equation_semantic/.style={draw=black,fill= none, minimum height=9mm, circle},
			equation_varnode/.style={draw=black,fill= none, text width=2.8mm, minimum height=5.5mm, circle},
			word/.style={draw=none,circle, minimum height=6mm, rectangle},
			chainLine/.style={line width=0.8pt,->, color=fontgrey}	%		,background rectangle/.style={fill=olive!45}, show background rectangle
			]
			%% Equation Tree
			\node[word](none) [] {};
			
			\node[equation_semantic](equ) [below = of none, yshift=6mm, xshift=73mm] {$+$};
			\node[equation_semantic](times) [below left = of equ, yshift=-4mm, xshift=-5mm] {$7$};
			\node[equation_semantic](addition) [below right = of equ, yshift=-4mm, xshift=5mm] {$+$};
			\node[equation_semantic](num1) [below left = of addition, xshift=0mm, yshift=-4mm] {$3$};
			\node[equation_semantic](var2) [below right = of addition, xshift=0mm, yshift=-4mm] {$6$};
			
			\draw [chainLine] (equ) to [] node[] {} (times);
			\draw [chainLine] (equ) to [] node[] {} (addition);
			\draw [chainLine] (addition) to [] node[] {} (num1);
			\draw [chainLine] (addition) to [] node[] {} (var2);
			
			\end{tikzpicture}
		} \\
		\textbf{Answer}\textcolor{white}{ioniion} $16$ \\
		%			\hline
		%			\textbf{Expression Tree}  \\
		%			%		\hline
		%			\scalebox{1.0}{\begin{tikzpicture}[node distance=2.0mm and 2.0mm, >=Stealth, 
		%				equation_semantic/.style={draw=black,fill= none, minimum height=9mm, circle},
		%				equation_varnode/.style={draw=black,fill= none, text width=2.8mm, minimum height=5.5mm, circle},
		%				word/.style={draw=none,circle, minimum height=6mm, rectangle},
		%				chainLine/.style={line width=0.8pt,->, color=fontgrey}	%		,background rectangle/.style={fill=olive!45}, show background rectangle
		%				]
		%				%% Equation Tree
		%				\node[word](none) [] {};
		%				
		%				\node[equation_semantic](equ) [below = of none, yshift=6mm, xshift=30mm] {$\times$};
		%				\node[equation_semantic](times) [below left = of equ, yshift=-4mm, xshift=-5mm] {$4.0$};
		%				\node[equation_semantic](addition) [below right = of equ, yshift=-4mm, xshift=5mm] {$+$};
		%				\node[equation_semantic](num1) [below left = of addition, xshift=0mm, yshift=-4mm] {$2.0$};
		%				\node[equation_semantic](var2) [below right = of addition, xshift=0mm, yshift=-4mm] {$6.0$};
		%				
		%				\draw [chainLine] (equ) to [] node[] {} (times);
		%				\draw [chainLine] (equ) to [] node[] {} (addition);
		%				\draw [chainLine] (addition) to [] node[] {} (num1);
		%				\draw [chainLine] (addition) to [] node[] {} (var2);
		%				
		%				\end{tikzpicture}}  
		\hline
		%			\hline
		\textbf{Problem 2} \\
		%					\hline
		{\em 3 times one of the numbers is 11 less than 5 times the other.}\\
		%		\hline 
		\textbf{Expression}\textcolor{white}{ionio}    $(3\times X_1)=(5\times X_2)-11$\\
		%			\textbf{Answer}\textcolor{white}{on} $X = 6.0$ \\
		%			\hline
		
		%		\hline
		\scalebox{0.7}{
			
			\begin{tikzpicture}[node distance=2.0mm and 2.0mm, >=Stealth, 
			equation_semantic/.style={draw=black,fill= none, minimum height=9mm, circle},
			equation_varnode/.style={draw=black,fill= none, text width=2.8mm, minimum height=5.5mm, circle},
			word/.style={draw=none,circle, minimum height=6mm, rectangle},
			chainLine/.style={line width=0.8pt,->, color=fontgrey}	%		,background rectangle/.style={fill=olive!45}, show background rectangle
			]
			%% Equation Tree
			\node[word](none) [] {};
			
			\node[equation_semantic](equ) [below = of none, yshift=6mm, xshift=70mm] {$=$};
			
			\node[equation_semantic](sub) [below right = of equ, yshift=-4mm, xshift=+10mm] {$-$};
			\node[equation_semantic](times2) [below left = of sub, yshift=-4mm, xshift=-3mm] {$\times$};
			\node[equation_semantic](num11) [below right = of sub, yshift=-4mm, xshift=3mm] {$11$};
			
			\node[equation_semantic](num5) [below left = of times2, xshift=0mm, yshift=-4mm] {$5$};
			\node[equation_semantic](var1) [below right = of times2, xshift=0mm, yshift=-4mm] {$X_2$};
			
			\node[equation_semantic](times1) [below left = of equ, yshift=-4mm, xshift=-10mm] {$\times$};
			\node[equation_semantic](num1) [below left = of times1, xshift=0mm, yshift=-4mm] {$3$};
			\node[equation_semantic](var2) [below right = of times1, xshift=0mm, yshift=-4mm] {$X_1$};
			
			\draw [chainLine] (equ) to [] node[] {} (sub);
			\draw [chainLine] (sub) to [] node[] {} (times2);
			\draw [chainLine] (times2) to [] node[] {} (num5);
			\draw [chainLine] (times2) to [] node[] {} (var1);
			\draw [chainLine] (sub) to [] node[] {} (num11);
			
			\draw [chainLine] (equ) to [] node[] {} (times1);
			\draw [chainLine] (times1) to [] node[] {} (num1);
			\draw [chainLine] (times1) to [] node[] {} (var2);
			\draw [chainLine] (times1) to [] node[] {} (var2);

			\end{tikzpicture}
		}  
		%			\\ 
		%			\hline
		
		%		\textbf{expression tree } \\
		%		\hline
		%			
		%		\\ \hline
\end{tabular}}
\vspace{0mm}
\caption{An example arithmetic word problem (top) and an example equation parsing problem (bottom) where the outputs can be represented as trees.}
\label{fig:examples}
\vspace{0mm}
\end{figure}
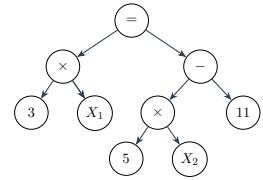

Designing computer algorithms that can automatically solve math word problems is a challenge for the AI research community \cite{bobrow1964natural}.
Two representative tasks have been proposed and studied recently -- solving arithmetic word problems \cite{wang2017deep,roy2018mapping,zou-19-qt} and equation parsing \cite{roy2016equation}, as illustrated in Figure \ref{fig:examples}.
The former task focuses on mapping the input paragraph (which may involve multiple sentences) into a target math expression, from which an answer can be calculated.
The latter task focuses on mapping a description (usually a single sentence) into a math equation that typically involves one or more unknowns.
As we can observe from Figure \ref{fig:examples}, in both cases, the output can be represented as a tree structure.

Earlier approaches to solving arithmetic word problems focused on rule-based methods where hand-crafted rules have been used \cite{mukherjee2008review,hosseini2014learning}.
Recently, learning-based approaches based on statistical classifiers \cite{kushman2014learning,roy2015solving,roy2016equation,liang2018meaning} or neural networks \cite{wang2017deep,wang2018mathdqn} have been used for making decisions in the expression\footnote{In this work, we use the term {\em expression} to refer to a math expression (for arithmetic word problem) or an equation (for equation parsing).} construction process.
However, these models do not focus on predicting the target tree as a complete structure at once, but locally trained classifiers are often used and local decisions are then combined.
Such local classifiers often make predictions on the choice of the underlying operator between two operands (e.g., numbers) appearing in the text in a particular order.
As a result, special treatments of the non-commutative operators such as {\em subtraction} ($-$) and {\em division} ($\div$) are often involved, where the introduction of inverse operators is typically required\footnote{For example, the inverse operator $-_{i}$ applied to two operands $a$ followed by $b$ is used to denote $b-a$.}.
%Research efforts in this area can date back to 1960s \cite{feigenbaum1963computers}.
%Solutions before 2014 most adopted rule-based approaches. 
%Recent progresses can be classified into five categories, as listed in Table \ref{tab:model_compare}: (1) rule-based methods \cite{mukherjee2008review,hosseini2014learning} apply a set of rules to map text to expressions;
%(A literature review \cite{mukherjee2008review} surveys most of rule-based methods before 2008).
%(2) statistics-based approaches \cite{kushman2014learning,roy2015solving,roy2016equation} make all related decisions via statistical classifiers;
%(3) DNN-based approaches translate the word sequences to math expressions via deep neural network models \cite{wang2017deep,wang2018mathdqn};
%(4) Meaning-based methods check the entity-attribute consistency between each quantity residing in the problem description and the goal of the question sentence \cite{liang2018meaning};
%(5) Semantic-based approaches \cite{shi2015automatically} address the math word problems by semantic parsing whose goal is to capture the underlying meaning of input text and map it to its corresponding meaning representation.
\citet{shi2015automatically} tackled the problem from a structured prediction perspective, where a semantic parsing algorithm using context-free grammars (CFG) was used.
However, their approach relies on semi-automatically generated rules and involves a manual step for interpreting the semantic representation they used.

While all these approaches focused on solving arithmetic word problems only, separate models have been developed for the task of equation parsing \cite{roy2016equation}. It is not clear how easy each of these models specifically designed for one task can be adapted for the other task.
Motivated by the observation that both problems involve mapping a text sequence to a tree structured representation, we propose \texttt{Text2Math} which regards both tasks as a class of structured prediction problems, and tackle them from a semantic parsing perspective.
We make use of an end-to-end latent-variable approach to automatically produce the target math expression at once as a complete structure, where no prior knowledge on the operators (such as whether an operator is non-commutative) is required. 
Our model outperforms all baselines on two benchmark datasets.
To the best of our knowledge, this is the first approach based on semantic parsing that tackles both arithmetic word problems and equation parsing with a single model.
Our code is available at  \url{http://statnlp.org/research/ta}.
%Unlike \cite{shi2015automatically}, we take a latent variable approach which is able to automatically acquire the rules from data used for performing semantic parsing.
%It assumes there exists a latent joint representation of both the text and the math expression tree that flexibly captures the correspondence between words and math expressions where node reordering can be explicitly modeled -- as a result, there is no need to define the inverse operators for non-commutative operators -- and involves minimal manual efforts in the learning process.

\section{Approach}
\label{sec:approach}

%In this section, we introduce our \texttt{Text2Math} model for semantically parsing text into math expressions.

\subsection{Expression Tree}
%Our goal is to design a semantic parsing approach for mapping text into the math expressions involved in the task of arithmetic math problems and equation parsing.
%As we have illustrated earlier, the math expressions for the tasks of arithmetic math problems and equation parsing can be represented with tree structures.
%Now let us formally define the target tree representations, which will then be regarded as the semantic representations used in the standard semantic parsing setup.

We first define tree representations for math expressions, which will then be regarded as the semantic representations used in the standard semantic parsing setup.

\begin{figure}[]
\centering
\scalebox{0.65}{
	\begin{tabular}{lm{6.5cm}}
		%			\hline	 
		{Math expression}   & $(7+(3+6))$ \\ 
		{Order in text}   & $(7, 3, 6) $  \\
		%\hdashline
		{Expression tree} &  
		\begin{tikzpicture}[node distance=1.8mm and 1.8mm, >=Stealth, 
		semantic/.style={draw=none,fill= none, text width=2.8mm, minimum height=5.5mm, circle},
		varnode/.style={draw=none,fill= none, text width=2.8mm, minimum height=5.5mm, circle},
		word/.style={draw=none,circle, minimum height=6mm, rectangle},
		chainLine/.style={line width=0.8pt,->, color=fontgrey}	%		,background rectangle/.style={fill=olive!45}, show background rectangle
		]
		%% Equation Tree
		\node[word](none) [] {};
		
		\node[semantic](add1) [below = of none, yshift=8mm, xshift=-2mm] {\textsc{Add}$_{(+)}$};
		\node[semantic](num29) [below left = of add1, yshift=0mm, xshift=-5mm] {\textsc{Con}$_{(7)}$};
		\node[semantic](add2) [below right = of add1, yshift=0mm, xshift=5mm] {\textsc{Add}$_{(+)}$};
		\node[semantic](num16) [below left = of add2, yshift=-4mm, xshift=-5mm] {\textsc{Con}$_{(3)}$};
		\node[semantic](num20) [below right = of add2, yshift=-4mm, xshift=5mm] {\textsc{Con}$_{(6)}$};
		
		\draw [chainLine] (add1) to [] node[] {} (num29);
		\draw [chainLine] (add1) to [] node[] {} (add2);
		\draw [chainLine] (add2) to [] node[] {} (num16);
		\draw [chainLine] (add2) to [] node[] {} (num20);
		
		\end{tikzpicture}  \\
		%		\hline
		\hline
		{Math expression} & $(3\times X_1)=(5\times X_2)-11$ \\ 
		{Order in text} & $(3, X_1, 11, 5, X_2)$ \\
		%\hdashline
		{Expression tree} 	&
		\begin{tikzpicture}[node distance=2mm and 1.8mm, >=Stealth, 
		semantic/.style={draw=none,fill= none, text width=2.8mm, minimum height=5.5mm, circle},
		varnode/.style={draw=none,fill= none, text width=2.8mm, minimum height=5.5mm, circle},
		word/.style={draw=none,circle, minimum height=6mm, rectangle},
		chainLine/.style={line width=0.8pt,->, color=fontgrey}	%		,background rectangle/.style={fill=olive!45}, show background rectangle
		]
		%% Equation Tree
		\node[word](none) [] {};
		
		\node[semantic](equ) [below = of none, yshift=8mm, xshift=-5mm] {\textsc{Equ}$_{(=)}$};
		\node[semantic](sub) [below right = of equ, yshift=0mm, xshift=5mm] {\textsc{Sub}$_{(-)}$};
		\node[semantic](mul2) [below left = of sub, xshift=-0mm, yshift=-4mm] {\textsc{Mul}$_{(\times)}$};
		\node[semantic](num5) [below left = of mul2, yshift=-4mm, xshift=-0mm] {\textsc{Con}$_{(5)}$};
		\node[semantic](var2) [below right = of mul2, yshift=-4mm, xshift=0mm] {\textsc{Var}$_{(X_2)}$};
		\node[semantic](num11) [below right = of sub, xshift=0mm, yshift=-4mm] {\textsc{Con}$_{(11)}$};
		
		\node[semantic](mul1) [below left = of equ, yshift=0mm, xshift=-10mm] {\textsc{Mul}$_{(\times)}$};
		\node[semantic](num3) [below left = of mul1, yshift=-4mm, xshift=-0mm] {\textsc{Con}$_{(3)}$};
		\node[semantic](var1) [below right = of mul1, yshift=-4mm, xshift=-2mm] {\textsc{Var}$_{(X_1)}$};

		\draw [chainLine] (equ) to [] node[] {} (sub);
		\draw [chainLine] (equ) to [] node[] {} (mul1);
		\draw [chainLine] (mul1) to [] node[] {} (num3);
		\draw [chainLine] (mul1) to [] node[] {} (var1);
		\draw [chainLine] (sub) to [] node[] {} (num11);
		\draw [chainLine] (sub) to [] node[] {} (mul2);
		\draw [chainLine] (mul2) to [] node[] {} (var2);
		\draw [chainLine] (mul2) to [] node[] {} (num5);
		
		\end{tikzpicture}  
		%			\\ 
		%			\hline
		
\end{tabular}}
%\vspace{-5mm}
\caption{Expression trees for the two math expressions in Figure \ref{fig:examples}. ``Order in text'' refers to the order that the textual expressions of operands appear in the problem text. We use subscripts to indicate the actual semantic interpretations.}
\label{fig:semantictree}
%\vspace{-5mm}
\end{figure}
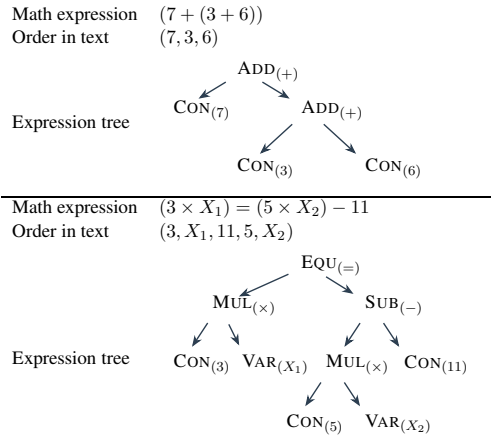

The nodes involved in the math expression trees can be classified into two categories, namely, {\em operator} and {\em quantity} nodes. 
Specifically, operator nodes are the tree nodes that define the types of operations involved in expressions.
In this work we consider \textsc{Add} ({\em addition}, $+$), \textsc{Sub} ({\em subtraction}, $-$), \textsc{Mul} ({\em multiplication}, $\times$) and \textsc{Div} ({\em division}, $\div$).
We also regard the equation sign ($=$) as an operation involved in math expressions and use \textsc{Equ} to denote it.
We consider two types of quantity nodes: \textsc{Con} denoting constants, and \textsc{Var} for unknown variables.
Table \ref{tab:semantic_unit} lists the above nodes.
Each tree node comes with an {\em arity} which specifies the number of direct child nodes that should appear below the given node. For example, the operator node \textsc{Sub} with arity 2 is expecting two child nodes below it in the expression tree, while \textsc{Con} with arity 0 is supposed to be a leaf node.
The two math expressions in Figure \ref{fig:examples} can be equivalently represented by expression trees consisting of such nodes, as illustrated in Figure \ref{fig:semantictree}.

\begin{table}[tp]
\centering
\scalebox{0.7}{
	\begin{tabular}{cllc}
		\hline
		Category & Node & Interpretation & Arity  \\
		\hline
		\multicolumn{1}{l}{\multirow{5}{*}{Operator}} &	\textsc{Equ}  & $n_i$ \textsc{Equ} $n_j$ $\Leftrightarrow$ $(n_i = n_j)$ &  2        \\
		&	\textsc{Add}  & $n_i$ \textsc{Add} $n_j$ $\Leftrightarrow$ $(n_i + n_j)$  & 2          \\
		&	\textsc{Sub}  &  $n_i$ \textsc{Sub} $n_j$ $\Leftrightarrow$ $(n_i - n_j)$  & 2    \\
		&	\textsc{Mul}  &   $n_i$ \textsc{Mul} $n_j$ $\Leftrightarrow$ $(n_i \times n_j)$&  2    \\
		&	\textsc{Div}  & $n_i$ \textsc{Div} $n_j$ $\Leftrightarrow$ $(n_i \div n_j)$ & 2  \\
		%			&	\textsc{Sub$_{R}$}  &  $n_i$ \textsc{Sub$_{R}$} $n_j$ $\Leftrightarrow$ $(n_j - n_i)$  & 2 \\
		%			&	\textsc{Div$_{R}$}  & $n_i$ \textsc{Div$_{R}$} $n_j$ $\Leftrightarrow$ $(n_j \div n_i)$  & 2 \\
		\hline
		\multicolumn{1}{l}{\multirow{2}{*}{Quantity}} & \textsc{Con}  & A constant & 0     \\
		&	\textsc{Var} & A variable & 0 \\
		\hline
\end{tabular}}
\vspace{0mm}
\caption{Expression tree nodes with interpretations, where $n_i$ ($n_j$) refers to the first (second) operand.}
\label{tab:semantic_unit}
\vspace{0mm}
\end{table}

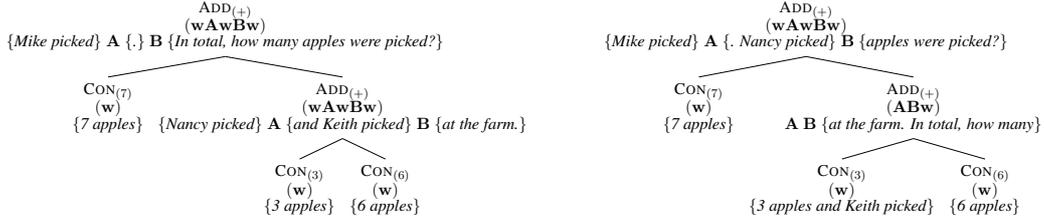
\begin{figure*}[tp]
	\centering
	\scalebox{0.75}{
		\begin{tabular}{cc}
			\multicolumn{2}{c}{{\em Mike picked 7 apples. Nancy picked 3 apples and Keith picked 6 apples at the farm. In total, how many apples were picked?}}     
			\\
			\\
			\scalebox{0.75}{
				\begin{tikzpicture}[node distance=1.5mm and 1mm, >=Stealth, 
				semantic/.style={draw=none,fill= none, text width=2.8mm, minimum height=5.5mm, rectangle},
				varnode/.style={draw=none,fill= none, text width=2.8mm, minimum height=5.5mm, circle},
				word/.style={draw=none,circle, minimum height=6mm, rectangle},
				chainLine/.style={line width=0.8pt,->, color=fontgrey}	%		,background rectangle/.style={fill=olive!45}, show background rectangle
				]
				%\node[text width = 1.5\textwidth] at (0,3.3) {\em Keith grew 29 cantelopes, Fred grew 16 cantelopes, and Jason grew 20 cantelopes. How many cantelopes did they grow in total?};
				
				\tikzset{align=center,level distance=55}
				\Tree	
				[ .{\textsc{Add}$_{(+)}$\\$(\mathbf{wAwBw})$\\\{{\em Mike picked\} $\mathbf{A}$ \{.\} $\mathbf{B}$ \{In total, how many apples were picked?}\}}
				[ .{\textsc{Con}$_{(7)}$\\$(\mathbf{w})$\\\{{\em 7 apples}\}}	]
				[ .{\textsc{Add}$_{(+)}$\\$(\mathbf{wAwBw})$\\\{{\em Nancy picked\} $\mathbf{A}$ \{and Keith picked\} $\mathbf{B}$ \{at the farm.}\}}
				[ .{\textsc{Con}$_{(3)}$\\$(\mathbf{w})$\\\{{\em 3 apples}\}}	
				]
				{\textsc{Con}$_{(6)}$\\$(\mathbf{w})$\\\{{\em 6 apples}\}}
				]	]
				
				%\node[word](v2)[xshift=-2mm, yshift=-60mm] {$(29+(16+20))$};
				
				%\node[single arrow, draw, align=center, xshift=0em, yshift=20mm, rotate=-90, minimum height=10mm](potok1){};
				%\node[single arrow, draw, align=center, xshift=0em, yshift=-48mm, rotate=-90, minimum height=10mm](potok2){};
				\end{tikzpicture}} 
			&
			\scalebox{0.75}{
				\begin{tikzpicture}[node distance=1.5mm and 1mm, >=Stealth, 
				semantic/.style={draw=none,fill= none, text width=2.8mm, minimum height=5.5mm, rectangle},
				varnode/.style={draw=none,fill= none, text width=2.8mm, minimum height=5.5mm, circle},
				word/.style={draw=none,circle, minimum height=6mm, rectangle},
				chainLine/.style={line width=0.8pt,->, color=fontgrey}	%		,background rectangle/.style={fill=olive!45}, show background rectangle
				]
				%\node[text width = 1.5\textwidth] at (0,3.3) {\em Keith grew 29 cantelopes, Fred grew 16 cantelopes, and Jason grew 20 cantelopes. How many cantelopes did they grow in total?};
				
				\tikzset{align=center,level distance=55}
				\Tree	
				[ .{\textsc{Add}$_{(+)}$\\$(\mathbf{wAwBw})$\\\{{\em Mike picked\} $\mathbf{A}$ \{. Nancy picked\} $\mathbf{B}$ \{apples were picked?}\}}
				[ .{\textsc{Con}$_{(7)}$\\$(\mathbf{w})$\\\{{\em 7 apples}\}}	]
				[ .{\textsc{Add}$_{(+)}$\\$(\mathbf{ABw})$\\{$\mathbf{A}$ $\mathbf{B}$ \{\em at the farm. In total, how many}\}}
				[ .{\textsc{Con}$_{(3)}$\\$(\mathbf{w})$\\\{{\em 3 apples and Keith picked}\}}	
				]
				{\textsc{Con}$_{(6)}$\\$(\mathbf{w})$\\\{{\em 6 apples}\}}
				]	]
				
				%\node[word](v2)[xshift=-2mm, yshift=-60mm] {$(29+(16+20))$};
				
				%\node[single arrow, draw, align=center, xshift=0em, yshift=20mm, rotate=-90, minimum height=10mm](potok1){};
				%\node[single arrow, draw, align=center, xshift=0em, yshift=-48mm, rotate=-90, minimum height=10mm](potok2){};
				\end{tikzpicture}} 
			\\

	\end{tabular}}
%	\vspace{-2mm}
	\caption{Example text-math trees for the arithmetic word problem example in Figure \ref{fig:examples}. The left tree captures the semantic correspondence well, while the right tree fails to capture the correct correspondence.}
	\label{fig:joint}
%	\vspace{-5mm}
\end{figure*}

\subsection{Latent Text-Math Tree}

With the specifically designed expression trees for representing the math expressions, we will now be able to design a model for parsing the text into the expression tree.
This is essentially a semantic parsing task.
One of the key assumptions made by the various semantic parsing algorithms is the intermediate joint representation used for connecting the words and semantics \cite{WongYW:06,zettlemoyer2007online:07,Luw:08,artzi2013weakly}.
\begin{comment}
.
For example, the work of \cite{WongYW:06} used a phrase-based machine translation based approach where the joint representation is assumed to be a derivation tree based on the learned synchronous context free grammar (SCFG) rules.
Similarly, Combinatory categorial grammar (CCG) derivation trees are used as the joint representation for the CCG-based semantic parsing algorithms \cite{zettlemoyer2007online:07,artzi2013weakly}.
\end{comment}
In this work, we adopt an approach that is inspired by \cite{Luw:08,Luw:14}, which learns a latent joint representation for words and semantics in the form of {\em hybrid trees} where word-semantics correspondence information is captured.
Specifically, we introduce a {\em text-math tree} representation that jointly encodes both text and the math expression tree.
Such joint representations can be understood as a modified expression tree where each semantic node is now augmented with additional word information from the corresponding text.

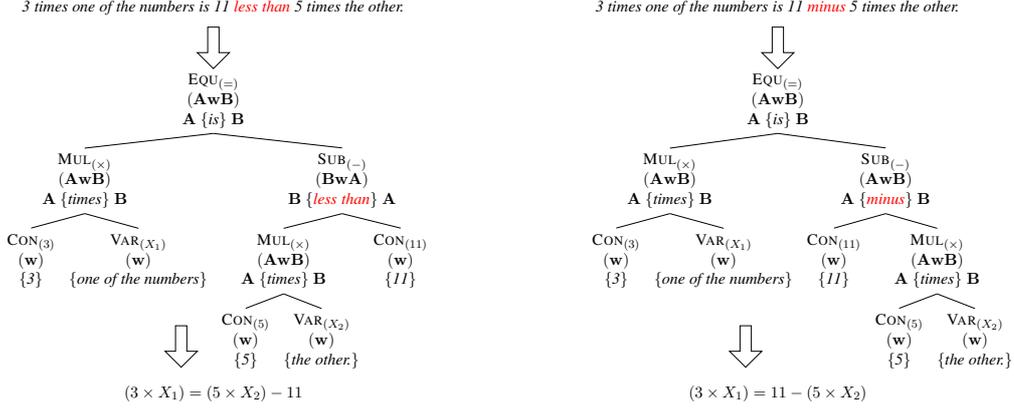
\begin{figure*}[h!]
	\begin{subfigure}{.47\textwidth}
		\centering
		\scalebox{0.55}{
			\begin{tikzpicture}[node distance=1.5mm and 1mm, >=Stealth, 
			semantic/.style={draw=none,fill= none, text width=2.8mm, minimum height=5.5mm, rectangle},
			varnode/.style={draw=none,fill= none, text width=2.8mm, minimum height=5.5mm, circle},
			word/.style={draw=none,circle, minimum height=6mm, rectangle},
			chainLine/.style={line width=0.8pt,->, color=fontgrey}	%		,background rectangle/.style={fill=olive!45}, show background rectangle
			]
			%% Equation Tree
			\node[word](v1)[xshift=0mm, yshift=28mm] {\em 3 times one of the numbers is 11 {\color{red}less than} 5 times the other.};
			
			\tikzset{align=center,level distance=55}
			\Tree	
			[ .{\textsc{Equ}$_{(=)}$\\$(\mathbf{AwB})$\\$\mathbf{A}$ \{{\em is}\} $\mathbf{B}$}
			[ .{\textsc{Mul}$_{(\times)}$\\$(\mathbf{AwB})$\\$\mathbf{A}$ \{{\em times}\} $\mathbf{B}$}
			[ .{\textsc{Con}$_{(3)}$\\$(\mathbf{w})$\\\{{\em 3}\}}	
			]
			[ .{\textsc{Var}$_{(X_1)}$\\$(\mathbf{w})$\\\{{\em one of the numbers}\}}
			]	]
			[ .{\textsc{Sub}$_{(-)}$\\$(\mathbf{BwA})$\\$\mathbf{B}$ \{{\color{red}{\em less than}}\} $\mathbf{A}$}
			[ .{\textsc{Mul}$_{(\times)}$\\$(\mathbf{AwB})$\\$\mathbf{A}$ \{{\em times}\} $\mathbf{B}$}
			{\textsc{Con}$_{(5)}$\\$(\mathbf{w})$\\\{{\em 5}\}}
			{\textsc{Var}$_{(X_2)}$\\$(\mathbf{w})$\\\{{\em the other.}\}}	
			] {\textsc{Con}$_{(11)}$\\$(\mathbf{w})$\\\{{\em 11}\}}		
			]
			]
			
			\node[word](v2)[xshift=0mm, yshift=-65mm] {$(3\times X_1)=(5\times X_2)-11$};
			
			\node[single arrow, draw, align=center, xshift=0em, yshift=19mm, rotate=-90, minimum height=10mm](potok1){};
			\node[single arrow, draw, align=center, xshift=-2em, yshift=-53mm, rotate=-90, minimum height=10mm](potok2){};
			\end{tikzpicture} }
		\label{fig:hybrid3}
	\end{subfigure}	\begin{subfigure}{.48\textwidth}
		\centering
		\scalebox{0.55}{
			\begin{tikzpicture}[node distance=1.5mm and 1mm, >=Stealth, 
			semantic/.style={draw=none,fill= none, text width=2.8mm, minimum height=5.5mm, rectangle},
			varnode/.style={draw=none,fill= none, text width=2.8mm, minimum height=5.5mm, circle},
			word/.style={draw=none,circle, minimum height=6mm, rectangle},
			chainLine/.style={line width=0.8pt,->, color=fontgrey}	%		,background rectangle/.style={fill=olive!45}, show background rectangle
			]
			%% Equation Tree
			\node[word](v1)[xshift=0mm, yshift=28mm] {\em 3 times one of the numbers is 11 {\color{red} minus} 5 times the other.};
			
			\tikzset{align=center,level distance=55}
			\Tree	
			[ .{\textsc{Equ}$_{(=)}$\\$(\mathbf{AwB})$\\$\mathbf{A}$ \{{\em is}\} $\mathbf{B}$}
			[ .{\textsc{Mul}$_{(\times)}$\\$(\mathbf{AwB})$\\$\mathbf{A}$ \{{\em times}\} $\mathbf{B}$}
			[ .{\textsc{Con}$_{(3)}$\\$(\mathbf{w})$\\\{{\em 3}\}}	
			]
			[ .{\textsc{Var}$_{(X_1)}$\\$(\mathbf{w})$\\\{{\em one of the numbers}\}}
			]	]
			[ .{\textsc{Sub}$_{(-)}$\\$(\mathbf{AwB})$\\$\mathbf{A}$ \{{\color{red} {\em minus}}\} $\mathbf{B}$}
			{\textsc{Con}$_{(11)}$\\$(\mathbf{w})$\\\{{\em 11}\}}
			[ .{\textsc{Mul}$_{(\times)}$\\$(\mathbf{AwB})$\\$\mathbf{A}$ \{{\em times}\} $\mathbf{B}$}
			{\textsc{Con}$_{(5)}$\\$(\mathbf{w})$\\\{{\em 5}\}}
			{\textsc{Var}$_{(X_2)}$\\$(\mathbf{w})$\\\{{\em the other.}\}}	
			]
			]	]
			
			\node[word](v2)[xshift=0mm, yshift=-65mm] {$(3\times X_1)=11-(5\times X_2)$};
			
			\node[single arrow, draw, align=center, xshift=0em, yshift=19mm, rotate=-90, minimum height=10mm](potok1){};
			\node[single arrow, draw, align=center, xshift=-2em, yshift=-53mm, rotate=-90, minimum height=10mm](potok2){};
			\end{tikzpicture} }
		\label{fig:hybrid4}
	\end{subfigure}
%	\vspace{-2mm}
	\caption{Left: example text-math tree for the equation parsing example in Figure \ref{fig:examples}, where the word association pattern $\mathbf{BwA}$ in the example is used for modeling reordering. Right: another example for a slightly different instance where the reordering is not required.}
	\label{fig:joint2}
	\vspace{-4mm}
\end{figure*}

From the joint representations we would be able to recover the semantic level correspondence information between words and math expressions.
%This joint representation is able to recover correct correspondence between words and expression tree nodes which will be helpful in the parsing process.
Possible joint representations of the two examples in Figure \ref{fig:examples} are illustrated in Figure \ref{fig:joint} and Figure \ref{fig:joint2} (left), respectively.
Consider Problem 1 in Figure \ref{fig:examples}. We illustrate two possible text-math expression trees in Figure \ref{fig:joint}.
Here each node in such joint representations is essentially a node in the original expression trees augmented with words from the problem text. For example, consider the left tree in Figure \ref{fig:joint}, the root node is an operator (the root node of the original expression tree) paired with discontiguous sequence of words ``\{{\em Mike picked}\}\dots\{{\em .}\}\dots\{{\em In total, how many apples were picked?}\}'' that appear in the problem text. This way, such a text-math tree is able to capture the semantic correspondence between words and basic units involved in the math expressions (i.e., operators, quantities).
%Consider the representation of Problem 2 in Figure \ref{fig:joint2}, {\color{red}LUWEI: Please change this discussion to something related to Figure 3 instead, as Figure 3 is on this page.... the root node \textsc{Equ} maps to the complete sentence $\boldsymbol{x}$, where the word sequence ``{\em is}" is immediately associated with it, while other words are collected from immediate associations of its children. Words ``{\em times}" and ``{\em less than}" are immediately associated to two expression tree nodes \textsc{Mul} and \textsc{Sub} respectively. The direct associations between word sequences and expression tree nodes are expected to reflect their semantic closeness, which is paramount knowledge for selecting arithmetic operators and makes generations interpretable.}
However, the text-math trees are not explicitly given during the training phase.
For example, the right side of Figure \ref{fig:joint} gives an alternative text-math tree that can also serve as a joint representation of both the text and the expression tree.
Comparing both trees we may see the one on the left appears to be better at capturing the true semantic level correspondence between words and math expressions.
Since there is no gold text-math tree explicitly given, we model it with a latent-variable approach.

Formally, given a text $\boldsymbol{x}$, paired with the expression $\boldsymbol{y}$ (or equivalently, the expression tree), we assume there exists a latent joint text-math representation in the form of text-math tree, that comprises exactly $\boldsymbol{x}$ and $\boldsymbol{y}$, denoted as $\boldsymbol{t}$.
%Such a joint representation is essentially a tree.
Each node is a word-semantics association $\left<x,y,p\right>$ where $x$ is a (possibly discontiguous) word sequence of $\boldsymbol{x}$ and $y$ is an individual expression tree node from $\boldsymbol{y}$, and $p$ is the word association pattern that is used to specify how words interact with the expression tree (further details will be provided in Sec. \ref{sec:inference}).
Intuitively, such a joint text-math representation should precisely contain the exact information associated with the text and its corresponding math expression and nothing else.
We will defer the discussion on how to exactly construct such joint representations until Sec. \ref{sec:inference}.

%Excluding the text information from such a joint representation, we arrive at the original math expression tree. On the other hand, the original text can be recovered from the yield of such a joint tree, with the help of word association patterns, to be discussed in the next section.

%For each node $(x,y, p)$, each word is only allowed to be associated with exactly one expression tree node.
%We regard words that appear at different positions in $\boldsymbol{x}$ as distinct words, regardless of their string forms.

The training corpus provides both the problem text $\boldsymbol{x}$ and its math expression, which we represent with an expression tree $\boldsymbol{y}$.
The joint representation $\boldsymbol{t}$ is not available in the training data, which we model as a latent variable.
The conditional random fields (CRF) \cite{lafferty2001conditional} has been successfully applied to many tasks in the NLP community \cite{lample2016neural,zou-lu-2018-learning,zou2019joint}.
In this work, we also apply CRF to model the conditional probability of the latent variable $\boldsymbol{t}$ and output expression $\boldsymbol{y}$, conditioned on the input $\boldsymbol{x}$.
The objective is defined as follows: 

\begin{gather}
P_{\Lambda,{\Theta}}(\boldsymbol{y}|\boldsymbol{x}) =\sum_{\boldsymbol{t}\in \mathcal{T}{(\boldsymbol{x},\boldsymbol{y})}} P_{\Lambda,{\Theta}}(\boldsymbol{y},\boldsymbol{t}|\boldsymbol{x})   \notag \\
=\frac{\sum_{\boldsymbol{t} \in \mathcal{T}{(\boldsymbol{x},\boldsymbol{y})}} e^{[\Lambda\cdot\Phi(\boldsymbol{x},\boldsymbol{y},\boldsymbol{t}) +G_{\Theta}(\boldsymbol{x},\boldsymbol{y},\boldsymbol{t})]}}{\sum_{\boldsymbol{y}',\boldsymbol{t}'\in \mathcal{T}(\boldsymbol{x},\boldsymbol{y}')}e^{[\Lambda\cdot\Phi(\boldsymbol{x},\boldsymbol{y}',\boldsymbol{t}')+G_{\Theta}(\boldsymbol{x},\boldsymbol{y}',\boldsymbol{t}')]}}
\label{eq:marginal}
\end{gather}
where $\Phi(\boldsymbol{x},\boldsymbol{y},\boldsymbol{t})$ returns a list of discrete features defined over the tuple $(\boldsymbol{x},\boldsymbol{y},\boldsymbol{t})$,
$\Lambda$ is the feature weight vector, $G_{\Theta}$ is a neural scoring function parameterized by $\Theta$ and
$\mathcal{T}(\boldsymbol{x}, \boldsymbol{y})$ is a set of  possible joint representations (i.e., text-math trees) for the pair $(\boldsymbol{x},\boldsymbol{y})$.

\subsection{Inference}
\label{sec:inference}

One challenge associated with the inference procedure in both training and decoding is how to handle the large space of latent structures defined by $\mathcal{T}(\boldsymbol{x,y})$ and $\mathcal{T}(\boldsymbol{x})$.
Without any constraints, searching or calculation that involves all possible structures within this space may be intractable.
We therefore introduce some assumptions on the set of allowable structures, such that tractable inference can be applied to such structures.

We first introduce three symbols $\mathbf{A}$, $\mathbf{B}$ and $\mathbf{w}$.
The symbol $\mathbf{A}$ refers to a placeholder for the left sub-tree (rooted by the left child node), and similarly $\mathbf{B}$ is a placeholder for the right sub-tree.
The symbol $\mathbf{w}$ refers to a contiguous sequence of (1 or more) words.
We will then use these three symbols to define the set of {\em word association patterns}, which are used to specify how the words interact with the sub-trees of the current node.
%allowable text-augmented expression trees.
Specifically, for expression tree nodes with arity 0 (i.e., quantity nodes), only one pattern $\mathbf{w}$ is allowed to be attached to them, indicating that a contiguous word sequence from a given problem text is associated with such expression tree nodes.
As for expression tree nodes with arity 2 (i.e., operator nodes), we define 16 allowable patterns denoted as $\{\mathbf{[w]A[w]B[w]}, \mathbf{[w]B[w]A[w]}\}$, where [] denotes optional.
Based on such word association patterns, we will be able to define a set of possible text-math trees for a particular text-expression pair that we regard as valid.
%Here, $\mathbf{A}$ and $\mathbf{B}$ refer to word sequences that the first and second child expression tree nodes will map to, respectively.

Before we formally define what is a valid text-math tree, let us look at an example in Figure \ref{fig:joint} which shows two valid trees.
First of all, we can verify that, if we exclude the words from both trees, we arrive at the math expression that corresponds to the text.
Second, we can also recover the text information from such a joint representation.
Let us look at the right tree in Figure \ref{fig:joint}.
Consider the right sub-tree of the complete tree rooted by the node $\left<x,y,p\right>=$ $\left<\right.$\{{\em at the farm. In total, how many}\}, \textsc{Add},  $\left.\mathbf{ABw}\right>$.
If we replace the placeholders $\mathbf{A}$ and $\mathbf{B}$ with the word sequences associated with its left and right sub-trees, respectively, we will arrive at the word sequence ``{\em 3 apples and Keith picked 6 apples at the farm. In total, how many}''.
Recursively performing such a rewriting procedure in a bottom-up manner, we will end up with a word sequence which is exactly the original input text as illustrated at the top of Figure \ref{fig:joint}.

Based on the above discussion, we can define $\mathcal{T}(\boldsymbol{x,y})$ as a set that consists of the valid trees:
\begin{definition}
For a given text $\boldsymbol{x}$ and an expression tree $\boldsymbol{y}$, a {\bf {\em valid text-math tree}} satisfies the following two properties:
1) the semantics portion of the tree gives exactly $\boldsymbol{y}$, and 2) the text obtained through the recursive rewriting procedure discussed above gives exactly $\boldsymbol{x}$\footnote{We regard words that appear at different positions in $\boldsymbol{x}$ as distinct words, regardless of their string forms.}.
\end{definition}

Given the definition of the valid text-math trees, we will be able to use a bottom-up procedure to construct the set $\mathcal{{T}(\boldsymbol{x},\boldsymbol{y})}$. Similarly, we will be able to construct the set $\mathcal{T}(\boldsymbol{x})$ by considering a forest-structured semantic representation that encodes all possible expression trees following \cite{Luw:15}. 
One nice property associated with considering only such joint representations is that there are known algorithms that can be used for performing efficient inference. Indeed, the resulting text-math trees are similar to the {\em hybrid tree} representations used in \cite{Luw:08,Luw:14}\footnote{They need to handle semantic nodes with arity 1 (which requires special constraints for properly defining $\mathcal{T}(\boldsymbol{x})$ \cite{Luw:15}), and their semantic nodes are also assumed to convey semantic type information for guiding the expression tree construction process, while we do not need to consider them.}, where dynamic programming based inference algorithms have been developed.
Such algorithms allow $\mathcal{O}(n^3m)$ time complexity for inference where $n$ is the text length and $m$ is the number of grammar rules\footnote{The grammars are related to the word association patterns. The possible latent text-math trees are constructed based on such grammar rules.} associated with the latent text-math trees.

We note that some prior systems \cite{roy2017unit,roy2018mapping} require extra inverse operators -- inverse subtraction ``$-_{r}$" and inverse division ``$\div_{r}$" to handle the scenarios where the order of quantities appearing in the text is not consistent with the order that they appear in the expression.
Exemplified by the example in the left of Figure \ref{fig:joint2}, by introducing two operators $-_{r}$ and $\div_{r}$ to take their operands in a reverse order, the equation on the left is represented as ``$(3\times X_1) = 11 -_{r} (5 \times X_2)$".
However, we do not need such two inverse operators.
The two group patterns $\mathbf{[w]A[w]B[w]}$ and $\mathbf{[w]B[w]A[w]}$ are capable to capture both orders.
A pattern from the first group handles the order that is consistent with the problem text, while a pattern from the second group is able to capture reordering of operands below an operator.
Exemplified by Figure \ref{fig:joint2}, reordering is required for the first example, but not for the second, though their texts only differ slightly.
Unlike the second example, instead of using the pattern $\mathbf{AwB}$, the first joint representation adopts the pattern $\mathbf{BwA}$ for the \textsc{Sub} expression node.
%, while $\mathbf{B}$ and $\mathbf{A}$ refer to two word sequences that directly map to the left and right child expression tree nodes.
Thus, our model is able to work without the underlying knowledge on whether an operator is commutative or not.

\subsection{Features}
\label{sec:feat}

\textbf{Discrete Features.} The feature function $\Phi(\boldsymbol{x},\boldsymbol{y},\boldsymbol{t})$ is defined over each node $\left<x,y,p\right>$ in the joint tree as well as the complete expression tree $\boldsymbol{y}$.
For each node $\left<x,y,p\right>$, we extract word n-gram, the word association pattern, and POS tags for words \cite{manning2014stanford}.
%We use the commonly-used Stanford Core NLP toolkit \cite{manning2014stanford} to preprocess the problem text and obtain part-of-speech (POS) tags for all words.
%Then POS tags are taken into account to capture syntactic information.
The knowledge that whether a number is relevant to the question (if available in the annotated data) is also taken as a binary feature.
%to alleviate the effects of irrelevant information.
%The relevance of each number, recognized by the Number Identifier, is also taken as a feature.
To assess the quality of the structure associated with the expression tree (i.e., features defined over $\boldsymbol{y}$), 
we extract parent-child relational information $(y_a, y_b)$ from $\boldsymbol{y}$, where $y_a$ is the parent of $y_b$, as features.
%Such features are analogous to the transition features used in linear-chain models such as first-order conditional random fields but are extracted from trees.
Following previous works \cite{roy2016equation,liang2018meaning}, we also consider incorporating a lexicon in our model so as to make a fair comparison with such works,
% to boost performance.
although we would like to stress that our model does not strictly require such lexicons for learning.
More details are in supplementary material.

\noindent
\textbf{Neural Features.} We design neural features over the pair of the $L$-sized window surrounding the target word $x_i$ in $\boldsymbol{x}$ and an expression tree node $y_j$.
The network takes as input the contiguous word sequence $(x_{i-L},\ldots, x_i,\ldots, x_{i+L})$, whose distributed representation is a simple concatenation of embeddings of each word.
The hidden layer applies an affine transformation with an element-wise nonlinear activation function, like tanh and ReLU.
The final output layer contains as many nodes as there are expression tree nodes in the training set.
The output is a score vector that gives a score for the input word sequence $(x_{i-L},\ldots, x_i,\ldots, x_{i+L})$ and an expression tree node $y_j$.
The neural scoring function is defined as follows:
%\begin{tabular}{l}
%	$G_{\Theta}(\boldsymbol{x},\boldsymbol{y},\boldsymbol{t}) =  \sum_{(x,y)\in\mathcal{W}(\boldsymbol{x},\boldsymbol{y},\boldsymbol{t})} c(x,y,\boldsymbol{x},\boldsymbol{y},\boldsymbol{t}) \times \psi(x,y)$
%\end{tabular}
\begin{gather}
G_{\Theta}(\boldsymbol{x},\boldsymbol{y},\boldsymbol{t}) = \notag \\ \sum_{(x,y)\in\mathcal{W}(\boldsymbol{x},\boldsymbol{y},\boldsymbol{t})} c(x,y,\boldsymbol{x},\boldsymbol{y},\boldsymbol{t}) \times \psi(x,y) \notag
\label{eq:neural_score}
\end{gather}
where $\mathcal{W}(\boldsymbol{x},\boldsymbol{y},\boldsymbol{t})$ is the set of $(x,y)$ pairs extracted from $(\boldsymbol{x},\boldsymbol{y},\boldsymbol{t})$, $c$ returns the number of occurrences and $\psi(x,y)$ is a score of the target word $x$ with $L$-sized windows and the expression tree node $y$, returned by the neural network.
We regard $L$ as a hyperparameter.

\subsection{Algorithms}

Given the complete training set, the log-likelihood is calculated as:
\begin{gather}
\mathcal{L}{(\Lambda, \Theta)}=\sum_{i}\log P_{\Lambda, \Theta}(\boldsymbol{y}_i|\boldsymbol{x}_i) %-\kappa||\Lambda||^2 
\notag \\   
=\sum_{i}\log \sum_{\boldsymbol{t}\in\mathcal{T}(\boldsymbol{x}_i,\boldsymbol{y}_i) }P_{\Lambda, \Theta}(\boldsymbol{y}_i,\boldsymbol{t}|\boldsymbol{x}_i)%-\kappa||\Lambda||^2
\label{eq:log}
\end{gather}
where $(\boldsymbol{x}_i$,$\boldsymbol{y}_i)$ refers to $i$-th instance in the training set.
The additional $L_2$ regularization term can be introduced to avoid over-fitting.
Here, we omit it for brevity.

The goal is to find optimal model parameters, i.e., $\Lambda$ and $\Theta$, which maximize the objective.
We first consider the computation of gradients for $\Lambda$. 
Assuming $\Lambda=\left<\lambda_1, \lambda_2, \ldots , \lambda_N\right>$, to learn the optimal feature weight values, we can calculate the gradient for each $\lambda_k$ in $\Lambda$ as:
\begin{gather}
\frac{\partial \mathcal{L}(\Lambda,\Theta)}{\partial{\lambda_{k}}} = \sum_i \sum_{\boldsymbol{t}} \boldsymbol{E}_{P_{\Lambda,\Theta}(\boldsymbol{t}|\boldsymbol{x}_i,\boldsymbol{y}_i)}[\phi_k(\boldsymbol{x}_i,\boldsymbol{y}_i,\boldsymbol{t})] \notag \\
-\sum_i \sum_{\boldsymbol{y},\boldsymbol{t}} \boldsymbol{E}_{P_{\Lambda,\Theta}(\boldsymbol{y},\boldsymbol{t}|\boldsymbol{x}_i)}[\phi_k(\boldsymbol{x}_i,\boldsymbol{y},\boldsymbol{t})]%-2\kappa\lambda_k
\label{eq:crf_grad}
\end{gather}
where $\phi_k(\boldsymbol{x},\boldsymbol{y},\boldsymbol{t})$ is the number of occurrences for the $k$-th feature extracted from $(\boldsymbol{x},\boldsymbol{y},\boldsymbol{t})$.

We then compute the gradient for the neural network parameters $\Theta$.
For an input word window $x$ and a semantic unit $y$, the gradient is defined as:
\begin{gather}
\frac{\partial \mathcal{L}(\Lambda,\Theta)}{\partial{\psi(x,y)}} = \sum_i \sum_{\boldsymbol{t}} \boldsymbol{E}_{P_{\Lambda,\Theta}(\boldsymbol{t}|\boldsymbol{x}_i,\boldsymbol{y}_i)}[c(x,y,\boldsymbol{x}_i, \boldsymbol{y}_i, \boldsymbol{t})] \notag \\
-\sum_i \sum_{\boldsymbol{y},\boldsymbol{t}} \boldsymbol{E}_{P_{\Lambda,\Theta}(\boldsymbol{y},\boldsymbol{t}|\boldsymbol{x}_i)}[c(x,y,\boldsymbol{x}_i, \boldsymbol{y}, \boldsymbol{t})]%-2\kappa\lambda_k
\label{eq:neural_grad}
\end{gather}

The gradients (\ref{eq:crf_grad},\ref{eq:neural_grad}) can be efficiently calculated by applying a generalized forward-backward algorithm, which allows us to conduct exact inference using the dynamic programming algorithm described in \cite{Luw:14}.
Next, standard methods like gradient descent, L-BFGS \cite{liu1989limited} can be used to find optimal values for model parameters.

%The objective (\ref{eq:log})  can be efficiently calculated by applying a generalized forward-backward algorithm, which allows us to conduct exact inference using the dynamic programming algorithm described in \cite{Luw:14}.
%Then, standard methods like L-BFGS \cite{Liu:89} can be used to find optimal values for model parameters.

During decoding, the optimal equation tree $\boldsymbol{y}^{\ast}$ for a new input $\boldsymbol{x}$ can be calculated by:
\begin{align}
\boldsymbol{y}^{\ast} &= \mathop{\arg\max}_{\boldsymbol{y}} P(\boldsymbol{y}|\boldsymbol{x}) \notag \\
&=\mathop{\arg\max}_{\boldsymbol{y}} \sum_{\boldsymbol{t}\in \mathcal{T}{(\boldsymbol{x},\boldsymbol{y})}}e^{F_{\Lambda,\Theta}{(\boldsymbol{x},\boldsymbol{y},\boldsymbol{t})}} \notag \\
&\approx \mathop{\arg\max}_{\boldsymbol{y},\boldsymbol{t}\in \mathcal{T}(\boldsymbol{x},\boldsymbol{y})}e^{F_{\Lambda,\Theta}{(\boldsymbol{x},\boldsymbol{y},\boldsymbol{t})}}
\label{eq:argmax}
\end{align}
where $\mathcal{T}(\boldsymbol{x}, \boldsymbol{y})$ refers to the set of all possible text-math trees that contain $\boldsymbol{x}$ and $\boldsymbol{y}$.

Instead of directly computing the summation over all possible latent text-math structures, we essentially replace the $\sum$ by the $\max$ operation inside the $\arg\max$.
In other words, we first find the latent text-math tree $\boldsymbol{t}^\ast$ which yields the highest score and contains the input text $\boldsymbol{x}$.
Then, the optimal expression tree $\boldsymbol{y}^\ast$ can be automatically extracted from $\boldsymbol{t}^\ast$.

An efficient dynamic programming based inference algorithm similar to the work of \citet{Luw:14} was leveraged to find the optimal latent structure $\boldsymbol{t}^\ast$.
We then obtain the optimal expression tree $\boldsymbol{y}^\ast$ from  $\boldsymbol{t}^\ast$,
which is the output of our system for the input problem text $\boldsymbol{x}$.

\subsection{Comparisons with \citet{roy2016equation}}
It is worth noting that \citet{roy2016equation} also proposed a system that maps text into an equation tree.
Unlike this work that maps math problem texts into math expressions in an end-to-end fashion, \citet{roy2016equation} designed three classifiers which sequentially make local decisions, namely identifying relevant numbers, recognizing possible variables and producing equations.
They also require extra inverse operators to handle the non-commutative operation issues, which is not necessary for our model.
Generating equations via a sequence of local classification decisions may propagate errors and even limit the ability to wholisticly understand the underlying semantics of problem texts which is important for predicting correct mathematical operations.
In this work, we regard equation parsing problem as a structure prediction task that allows to parse the text to equations from a semantic parsing perspective.
Moreover, \texttt{Text2Math} is capable to handle both tasks of equation parsing and arithmetic word problems, while the system of \citet{roy2016equation} is specific to equation parsing.

\section{Experiments}
\label{sec:exp}

\textbf{Datasets.} Following prior works \cite{roy2015solving,liang2018meaning}, we focus on two commonly-used benchmark datasets for arithmetic word problems, AI2 \cite{hosseini2014learning} and IL \cite{roy2015solving}.
We consider mathematical relations among numbers and calculate numerical values of the predicted expressions.
For equation parsing, we also evaluate our model on the data released by \cite{roy2016equation}.
A predicted equation is regarded as a correct one if it is mathematically equivalent to the gold equation.
%In this corpus, each instance can contain a maximum two variables mapped to noun phrases in the text.
%\noindent
%\textbf{Implementation Details}
%Our system is implemented in Java.
%For all of our experiments, we applied the L-BFGS algorithm \cite{liu1989limited} for learning model parameters.
%The maximum number of L-BFGS steps was set to 100.

\begin{table}[t]
\centering
\scalebox{0.68}{
	\begin{tabular}{ll|c|c|c}
		%		\hline
		\multicolumn{2}{l|}{System}& AI2 & IL & Average \\ 
		\hline
		\multicolumn{2}{l|}{$\ast$\citet{liang2018meaning} (Statistical)} & 81.5 & 81.0 & 81.25 \\
		\multicolumn{2}{l|}{$\ast$\citet{liang2018meaning} (DNN)} & 69.8 & 70.6 & 70.20 \\
		\multicolumn{2}{l|}{$\ast$\citet{roy2017unit}}  & 76.2 & 71.0 & 73.60 \\ %(UnitDep)
		\multicolumn{2}{l|}{\citet{roy2015solving}}  & 78.0 & 73.9 & 75.95 \\ %(ExpTree)
		\multicolumn{2}{l|}{\citet{koncel2015parsing}}  & 52.4 & 72.9 & 62.65 \\ %(ALGES)
		\multicolumn{2}{l|}{\citet{roy2015reasoning}}   & - & 52.7 & - \\
		\multicolumn{2}{l|}{\citet{kushman2014learning}}  & 64.0 & 73.7 & 68.85 \\ %(KAZB)
		\multicolumn{2}{l|}{\citet{hosseini2014learning}} & 77.7 & - & - \\
		\hline
		\hline
		\multicolumn{1}{p{2cm}}{\multirow{4}{*}{\textsc{Non-Neural}}}&\texttt{Text2Math} & {85.8} & 80.4 & 83.10\\
		&\textcolor{white}{12}\textsc{-pos} & \textbf{86.0} & \textbf{81.0} & \textbf{83.50}\\
&\textcolor{white}{12}\textsc{-lex}& 76.8 & 69.4 & 73.10  \\
		&\textcolor{white}{12}\textsc{-id} & 75.4 & 78.1& 76.75\\
\hdashline

\multicolumn{1}{p{2cm}}{\multirow{7}{*}{\textsc{Neural}}}&$L=0$& 84.8 & 79.7 & 82.25\\

& $L=1$& 84.5  & 80.3  & 82.40   \\
& $L=2$&85.5   & 80.3 & 82.90   \\
& $L=3$& {86.2}  & 80.9 &  {83.55}   \\
	& $L=4$& 85.5  &  80.0 & 82.75   \\
& $L=5$& {85.2}  & \textbf{81.4}   & 83.30  \\
& $L=6$& \textbf{86.5} &81.0	&\textbf{83.75}  \\
	\end{tabular}
}
\vspace{-2mm}
\caption{\textbf{Arithmetic Word Problem}: Accuracy (\%) on the two benchmark datasets. (\textsc{-pos}, \textsc{-lex}, \textsc{-id} mean that the model excludes POS tags feature, lexicon, or the number relevance feature.) $\ast$ indicates model uses prior knowledge, such as lexcion and inference rules.}
\label{tab:accuracy}
\vspace{-5mm}
\end{table}

\subsection{Empirical Results}
\textbf{Arithmetic Word Problem.} Following previous work \cite{liang2018meaning}, we conduct 3-fold and 5-fold cross-validation on AI2 and IL, respectively, and report the accuracy scores, as shown in Table \ref{tab:accuracy}.
Our method achieves competitive results on AI2 and IL.
Overall, it performs better than previous systems in terms of average scores.

Ablation tests have been done to investigate the effectiveness of different components, such as POS tags, the lexicon and number relevance, as indicated by ``\textsc{-pos}", ``\textsc{-lex}", ``\textsc{-id}" in Table \ref{tab:accuracy}.
By eliminating POS tag features, we achieve new state-of-the-art results on two datasets, which shows POS tag features do not appear to be helpful in this case.
Without using lexicon, the performance drops a lot as expected, but the results are still comparable with most previous systems.
These figures demonstrate the effectiveness of the lexicon.
It is worth noting that the work of \citet{liang2018meaning} that achieve previous state-of-the-art results leverage inference rules during the inference phase. Their approach can be regarded as a different way of using the lexicon similar to ours.
%It is aware that the generated POS tags are not perfect \cite{manning2014stanford}.
%One possible reason resulting in such a behavior is that data distributions and errors made by the tag generator of two corpora are different.
%The high precision lexicon does boost the performance for such two datasets, especially for the IL dataset.
%Such a dataset contains only single-step problems covered all four basic math operations, addition, subtraction, multiplication and division.
%The lexicon can accurately predict the type of a problem is such a problem meets one entry of the lexicon.
%Considering the relevance of numbers is necessary since not each number appearing in the problem text is relevant to the question.
We also consider the effects of neural features (see Sec. \ref{sec:feat}) with different window sizes $L\in\{0,1,2,3,4,5,6\}$.
According to empirical results, a larger window size tends to give better results.
One possible reason is that an arithmetic word problem often consists of several sentences, where a large word window is required to capture mathematical semantics.

%
%\begin{table}[tp]
%	\centering
%	\scalebox{0.9}{	
%		\begin{tabular}{ll|c}
%			\multicolumn{2}{l|}{System}     &Equation   \\	
%			\hline 
%			\multicolumn{2}{l|}{$\ast$\citet{roy2016equation} (Pipeline) }        & 71.3           \\		\hline
%			\multicolumn{2}{l|}{$\ast$\citet{roy2016equation} (Joint) }          &  60.9            \\
%			%			\hline 		
%			%			$\ast$ SPF \cite{artzi2013uw}         &  \textcolor{white}{0}3.1             \\
%			%			\hline
%			%		\textsc{Parser}-no num   &                      &           \\
%			\hline
%			\hline
%			\multicolumn{2}{l|}{\texttt{Text2Math}} &    71.4 \\
%			\multicolumn{2}{l|}{\textcolor{white}{12}\textsc{-pos}}  & 69.1     \\
%			\multicolumn{2}{l|}{ \textcolor{white}{12}\textsc{-lex}} &    71.4 \\
%			\hdashline
%			\multicolumn{2}{l|}{\textcolor{white}{12}+\textsc{g}}  &       \textbf{73.2}                 \\
%			\multicolumn{2}{l|}{\textcolor{white}{12}+\textsc{g-lex}}  &       \textbf{73.2}                 \\
%			
%			
%			%			\hline
%			%			\hline
%			\hdashline	
%			%					&$L=0$& 71.4  \\
%			%					& $L=1$& 71.9       \\
%			& $L=2$& 73.8       \\
%			%						& $L=3$& 73.5       \\
%			\multicolumn{1}{l}{\multirow{1}{*}{\textcolor{white}{12}+\textsc{Neural}}} & $L=4$& \textbf{74.5}  \\
%			%						& $L=5$& 74.0 \\
%			& $L=6$& 73.2 \\
%	\end{tabular}}
%	\vspace{-2mm}
%	\caption{\textbf{Equation Parsing}: Accuracy on equation parsing dataset. -\textsc{pos}: without POS tag features; -\textsc{lex}: without Lexicon; \textsc{+g}: with gold identification for numbers. 
%		$\ast$ indicates model uses lexicon.
%	}
%	\label{tab:equation}
%	\vspace{-5mm}
%\end{table}

\begin{table}[tp]
\centering
\scalebox{0.68}{	
	\begin{tabular}{ll|c}
		\multicolumn{2}{l|}{System}     &Equation   \\	
		\hline 
		\multicolumn{2}{l|}{$\ast$\citet{roy2016equation} (Pipeline) }        & 71.3           \\		\hline
		\multicolumn{2}{l|}{$\ast$\citet{roy2016equation} (Joint) }          &  60.9            \\
					 		
		\multicolumn{2}{l|}{$\ast$ SPF \cite{artzi2013uw}}         &  \textcolor{white}{0}3.1             \\
		%			\hline
		%		\textsc{Parser}-no num   &                      &           \\
		\hline
		\hline
		\multicolumn{1}{l}{\multirow{5}{*}{\textsc{Non-Neural}}}&\multicolumn{1}{l|}{\texttt{Text2Math}} &    71.4 \\
		&\multicolumn{1}{l|}{\textcolor{white}{12}\textsc{-pos}}  & 69.1     \\
&\multicolumn{1}{l|}{ \textcolor{white}{12}\textsc{-lex}} &    71.4 \\
		&\multicolumn{1}{l|}{\textcolor{white}{12}+\textsc{g}}  &       \textbf{73.2}                 \\
		&\multicolumn{1}{l|}{\textcolor{white}{12}+\textsc{g-lex}}  &       \textbf{73.2}                 \\
\hdashline	
		&$L=0$& 71.4  \\
		& $L=1$& 71.9       \\
& $L=2$& 73.8       \\
\multicolumn{1}{l}{\multirow{1}{*}{\textsc{Neural}}}			& $L=3$& 73.5       \\
 & $L=4$& \textbf{74.5}  \\
			& $L=5$& 74.0 \\
& $L=6$& 73.2 \\
\end{tabular}}
\vspace{-2mm}
\caption{\textbf{Equation Parsing}: Accuracy (\%) on equation parsing dataset. -\textsc{pos}: without POS tag features; -\textsc{lex}: without Lexicon; \textsc{+g}: with gold identification for numbers. 
	$\ast$ indicates model uses lexicon. Result of SPF \cite{artzi2013uw} is taken from \citet{roy2016equation}.
}
\label{tab:equation}
\vspace{-6.5mm}
\end{table}

\noindent
\textbf{Equation Parsing.} We compare our model with previous work \cite{roy2016equation} on the equation parsing dataset, as shown in Table \ref{tab:equation}.
%Unlike systems of \cite{roy2016equation} (\textbf{Pipeline} and \textbf{Joint}) and \cite{artzi2013uw} (\textbf{SPF}), an external lexicon specific for mathematical expressions is not required for this task.
Our method yields competitive results.
Unlike the work of \citet{roy2016equation}, annotations of unknown variables are not required in our model.
As reported in \cite{roy2016equation}, 
they trained SPF \cite{artzi2013uw}, a publicly available semantic parser, with sentence-equation pairs and a seed lexicon for mathematical terms.
But it only obtained $3.1\%$ accuracy.
The result taken from \cite{roy2016equation} shows that it might be difficult for such a semantic parser in handling the equation parsing task even with a high precision lexicon.
One possible reason is that mapping text into a math equation is essentially a structure prediction problem.
Our model is capable to make guaranteed decisions from a structure prediction perspective.
Different from arithmetic word problems, where numbers are explicitly given in the form of digits, some texts from equation parsing corpus describe numbers in string forms.
Hence, a structured predictor is used to identify the numbers in the sentence, which achieves $95.3\%$ accuracy.
The identifications of numbers are taken as features.
We also consider the gold label of numbers, indicated by \textsc{(+g)}.
%To investigate how the identification of relevant numbers affects the prediction of equations, we use the gold 
The performance improves a lot, which shows that the accurate identification of numbers is necessary to in order to obtain a good performance.
By removing POS tag features, there is a slight drop in accuracy.
On the other hand, it is worth noting that even without the high precision lexicon, our model can still achieve new state-of-the-art accuracy in this task, while the previous work \cite{roy2016equation} always requires a high precision lexicon to boost performance.
Incorporating neural features leads to new state-of-the-art accuracy of 74.5\% when $L=4$.

\noindent
\textbf{Expression Construction.}
In arithmetic word problems, the expression consists of several numbers only, exemplified by Problem 1 in Figure \ref{fig:examples}.
In practice, an unknown variable X, representing the goal that the problem aims to calculate, can be appended to the expression to form an equation.
We further investigate two constructions: appending the unknown variable X to the beginning or to the end of an expression.
Results are listed in the first block of Table \ref{tab:reverse}.
For instance, two new constructions of the running example are $X=(29+(16+20))$ as indicated by ``Prefix X", and $(29+(16+20))=X$ reported as ``Suffix X".
%During training, the word association of the unknown $X$ is learned where we did not constrain $X$ to be hard associated with the question sentence.
It is interesting that including an $X$ and its position influences the performance.
Overall, excluding $X$ works the best which is adopted in this work.
\begin{table}[t]
\centering
\scalebox{0.7}{
	\begin{tabular}{l|c|c|c}
		Variants & AI2 & IL & Average\\ \hline
		\texttt{Text2Math} & 85.8 & 80.4 & 83.10 \\ 
		Prefix X & 84.3 & 81.0 & 82.65 \\ 
		Suffix X & 84.3 & \textbf{81.2} & 82.75 \\ \hdashline
		\texttt{Text2Math} + Inverse & \textbf{86.3} & 80.1 & \textbf{83.20} \\ 
		Prefix X + Inverse & 84.0 & 79.9 & 80.45\\ 
		Suffix X + Inverse & 84.0 & 80.5 & 82.25 \\
	\end{tabular}
}
\vspace{-2mm}
\caption{Performance of different constructions for expression of arithmetic word problems and the effects of incorporating inverse operators.}
\label{tab:reverse}
% \vspace{-5mm}
\end{table}

\noindent
\textbf{Inverse Operators.}
As we discussed in Sec. \ref{sec:inference}, one distinct advantage of our approach, as compared to others, is that we do not need inverse operators, such as ``$-_{r}$" and ``$\div_{r}$".
Our designed word association patterns are capable to handle the reordering issue.
Here, we consider model variants by introducing two inverse expression tree nodes, \textsc{Sub}$_{r}$ and \textsc{Div}$_{r}$, to represent ``$-_{r}$" and ``$\div_{r}$", respectively.
Empirical results, reported in the second block in Table \ref{tab:reverse}, show that \texttt{Text2Math} (without including inverse operators) can obtain comparative results compared to the model variants with inverse operators.
%However, it will slow down the training of our algorithm, since the search space becomes larger.
These results confirm that our model does not require additional knowledge of the semantics of operands, which is a unique property of our approach.

%We observe that the model with larger window size tends to give more guaranteed predictions for both arithmetic word problem solving and equation parsing task.
%The larger word window might be capable to capture more contextual information that is potentially helpful to disambiguate semantic meanings of the target word.
%On the other hand, the value of $L$ should be set properly according to the distribution of dataset. 
%It is not always the case that larger $L$ would lead to a better performance.

%Comparing predictions made by Pipeline \cite{roy2016equation} and our technique, we found that \textsc{Parser} shows its power of understanding the meaning of an input sentence.
%We illustrate two examples in Table \ref{tab:comparison}.
%The Pipeline fails capturing the meaning of ``{\em more than twice as many games as}" which implies addition and multiplication operations , while the \textsc{Parser} is capable to capture such knowledge.
%In the second example, our model can also map the phrase ``{\em 10 identical sweaters}" into its corresponding math expression $10\times X_1$.
%Moreover, \textbf{\textsc{Parser}} predicts more promising groundings of variables.

\subsection{Qualitative Analysis}
\textbf{Output Comparisons.} Equation parsing is more challenging than arithmetic word problems, since it requires generating unknown variables mapped to phrases residing in the text.
%Such a task demands understanding of the problem descriptions.
We analyze output of this task to investigate the source that leads to better performance.
Comparing predictions made by Pipeline \cite{roy2016equation} and our approach, we found that \texttt{Text2Math} can better capture the meaning of the problem text.
We illustrate two examples in Table \ref{tab:comparison}.
The Pipeline approach fails to capture the meaning of ``{\em rises to 36\% from 3.4\%}" which implies subtraction of two numbers, while our model is capable to capture such knowledge.
In the second example, Pipeline misunderstands the meaning of ``{\em five more than three}", although it seems correct in a local context.
However, an equation should be mapped from the complete sentence that captures mathematical relations in a global perspective.
Our model holds such a capability and makes more guaranteed predictions, which proves the efficacy of solving math problems from a structure prediction perspective.
\begin{table}[tp]
\centering
\scalebox{0.7}{
	\begin{tabular}{|lp{6.5cm}|}
		\hline
		\textbf{Input}: &{\em Japan January jobless rate rises to $3.6\%$ from $3.4\%$.} \\
		\textbf{Gold}: &$X_1 = 0.036-0.034$ \\ \hdashline
		\textbf{Pipeline}: &$X_1\times 0.036 = 0.034$ \\
		\texttt{Text2Math}: &$X_1 = 0.036-0.034$ \\
		%			\hline
		%			\textbf{Input}: &{\em The Mainers won 24 more than twice as many games as they lost.} \\
		%			\hdashline
		%			\textbf{Gold}: &$X_1=24+(2\times X_2)$ \\ 
		%			\textbf{Pipeline}: &$24=X_2/(X_1/2)$ \\
		%			\textbf{\textsc{Jr}}: & $X_1=24+(2\times X_2)$\\
		\hline
		\hline
		\textbf{Input}: &{\em The number of baseball cards he has is five more than three times the number of football cards.} \\
		\textbf{Gold}: &$X_1=5+(3\times X_2)$ \\
		\hdashline
		\textbf{Pipeline}: &$X_1\times X_2 =5-3$
		\\
		\texttt{Text2Math}: &$X_1=5+(3\times X_2)$
		\\
		\hline
		
\end{tabular}}
\vspace{-2mm}
\caption{Comparison between predictions made by the previous state-of-the-art system \cite{roy2016equation} (denoted as Pipeline) and \texttt{Text2Math}.}
\label{tab:comparison}
\vspace{-3mm}
\end{table}

\noindent
\textbf{Robustness.}
To further investigate the property of our model, we studied outputs.
We found that our method is able to conduct self-correction.
Exemplified by Example 3 in Table \ref{tab:error}, considering the sentence {\em ``Germany's DAX opens 0.7\% lower at 18,842."} with annotated equation $X_1+(0.007\times X_1)=18,842$, the prediction made by our method is $X_1-(0.007 \times X_1)=18,842$.
It can be seen that the prediction made by our method is supposed to be the correct one, while the annotation is actually wrong.
To make a fair comparison with previous works, we did not count such cases as correct during evaluation, which implies that accuracy reported in Table \ref{tab:equation} is in fact higher.
%On the other hand, this demonstrates that \textsc{Parser} is a robust system that is able to spot incorrect annotations.

\noindent
\textbf{Error Analysis.}
For arithmetic word problem, it is interesting that the operand of two operands should be addition/subtraction (multiplication/division), while the prediction is subtraction/addition (division/multiplication).
Consider Example 4 and 5 in Table \ref{tab:error}. Descriptions of such two problems share many words, such as {\em each}, {\em how many}, {\em there are}, etc.
Slight difference in problem descriptions may lead to different results, which makes it a challenge.

\begin{table}[tp]
\centering
\scalebox{0.75}{
	\begin{tabular}{|lp{7cm}|}
		\hline
		\textbf{Example 3}:& {\em Germany's DAX opens 0.7\% lower at 10,842.} \\
		\hdashline
		\textbf{Gold}:& $X_1+(0.007\times X_1)=10842$\\ \texttt{Text2Math}:& $X_1-(0.007\times X_1)=10842$\\
		\hline          
		\hline
		\textbf{Example 4}:& {\em Each child has 5 bottle caps. If there are 9 children, how many bottle caps are there in total?
		} \\
		\hdashline
		\textbf{Gold}:& $5\times 9$\\ 
		\texttt{Text2Math}:& $5 \div 9$\\
		\hline
		\hline
		\textbf{Example 5}:& {\em The school is planning a field trip. There are 14 students and 2 seats on each school bus. How many buses are needed to take the trip?
		} \\
		\hdashline
		\textbf{Gold}:& $14 \div 2$\\ 
		\texttt{Text2Math}:& $14 \times 2$\\
		\hline
		\hline
		\textbf{Example 6}:& {\em 530 pesos can buy 4 kilograms of fish and 2 kilograms of pork.} \\
		\hdashline
		\textbf{Gold}:& $ 530=(4 \times X_1)+(2\times X_2)$\\ 
		\texttt{Text2Math}:& $530\times X_3=(4\times X_1)+(2\times X_2)$\\
		\hline
		\hline
		\textbf{Example 7}:& {\em Flying with the wind , a bird was able to make 150 kilometers per hour.} \\
		\hdashline
		\textbf{Gold}:& $X_1+X_2=150
		$\\ 
		\texttt{Text2Math}:& $X_1=150$\\
		\hline
\end{tabular}}
% \vspace{-2mm}
\caption{Examples with wrong predictions. Gold denotes the annotated correct equations and \texttt{Text2Math} refers to output equations generated by our method.}
\label{tab:error}
\vspace{-3mm}
\end{table}

As for equation parsing, the work of \citet{roy2016equation} requires annotations on which phrases should be mapped to unknowns during the training phase. 
However, such supervised knowledge is not required for our method.
In our setup, we did not make hard constraint that each prediction must contain one or two variables.
Therefore, missing or redundant variables appearing in the predicted equations are one of the major error sources.
Example 6 and 7 from Table \ref{tab:error} illustrate such cases.
On the other hand, lack of professional background information also leads to missing variables.
Consider Example 6. Without world knowledge, it might be difficult for the algorithm to recognize that ``{\em Flying with the wind}" implies the speed of the wind which should be considered as a variable of the equation.
%So the understanding of time, distance, {\em etc}.

\section{Related Work}

%We briefly introduce recent research progresses which are related to this work.
%There are two major strands of research in NLP committee - arithmetic word problem solving and semantic parsing.

\noindent
\textbf{Math Word Problems.} 
%An arithmetic word problem typically consists of several sentences describing a math problem.
%The text contains at least two numbers and implies basic mathematical operations among them.
%The goal of automatically solving arithmetic word problems is to reveal the underlying mathematical relations among numbers by understanding the problem text.
%Solutions before 2014 all applied rule-based approach.
\citet{mukherjee2008review} surveyed related approaches to this task in literature.
%Recently, researchers made progresses with various learning algorithms designed to automatically solve arithmetic word problems.
\begin{comment}
	\citet{hosseini2014learning} observed that verbs in the text describe the impact on the world state and thus mapped verbs into seven defined categories, each of which implied a mathematical operation.
	\citet{mitra2016learning} categorized problems into three types, i.e., \emph{part-whole, change} and \emph{comparison}
	%Each type is associated with a specific math formula, which is used to fill slot with numbers and solve problems. 
	to solve simple addition-subtraction problems.
\end{comment} 
\citet{hosseini2014learning,mitra2016learning} solved the task by categorizing verbs or problems.
The first method that can handle general arithmetic problems with multiple steps was proposed by \citet{roy2015solving}, which was further extended by introducing \cite{roy2017unit,roy2018mapping}.
%The same researchers further proposed different strategies to make automatic solvers more promising and generalizable 
\citet{zou-19-qt,zou-19-qt-supp} is the first work that proposed a sequence labelling approach to solving arithmetic word problems, which focuses on addition-subtraction word problems.
Other systems include semantic parsing based approaches \cite{liang2018meaning} and neural methods \cite{wang2017deep, wang2018translating,wang2018mathdqn}.
Unlike arithmetic word problems, the goal of algebra word problems is to map the text to an equation set \cite{kushman2014learning,shi2015automatically}.
Other types of problems have also been investigated, including probability problems \cite{dries2017solving}, logic puzzle problems \cite{mitra2015learning,chesani2017solving} and geometry problems \cite{seo2014diagram,seo2015solving}.
Besides the benchmark datasets used in this work, other popular datasets include Dolphin18K \cite{shi2015automatically} and AQuA \cite{ling2017program} for algebra word problems which are not the focus in this work.
\citet{roy2016equation} first proposed the Equation Parsing task and designed a pipeline method with three structured predictors.
%We focus on generating equations for now and leave grounding for future work.

\noindent
\textbf{Semantic Parsing.}
Another line of related works is semantic parsing \cite{WongYW:06,zettlemoyer2007online:07,kwiatkowski2010inducing:10,liang2011learning:11,P18-1068,zou-lu-2018-learning}, which aims to map sentences into logic forms,
%These techniques are typically under full supervision or semi-supervision, where sentences paired with annotated logic forms are available during training phrase.
including CCG-based lambda calculus expressions \cite{zettlemoyer2007online:07,kwiatkowski2010inducing:10,artzi2013weakly,P16-1004},
FunQL \cite{kate2005learning,WongYW:06,Jones:12}, lambda-DCS \cite{liang2011learning:11,berant2013semantic:13,jia2016data}, graph queries \cite{harris2013sparql,holzschuher2013performance} and SQL \cite{yin2015neural,sun2018semantic}.
In this work, we adopt a text-math semantic representation encoding words and the expression tree.
%In this work, we aim to understand the problem text and map it to a math expression which recovers mathematical relations among numbers.
%The training corpus often consists of problems texts associated corresponding equations and answers.

%In this work, we introduce a semantic parsing algorithm that allows us to map an arithmetic word problem to a math expression or an equation.

\section{Conclusion}
In this work, we propose a unified structured prediction approach, \texttt{Text2Math}, to solving both arithmetic word problems and equation parsing tasks.
We leverage a novel joint representation to automatically learn the correspondence between words and math expressions which reflects semantic closeness.
Different from many existing models, \texttt{Text2Math} is agnostic of the semantics of operands and learns to map from text to math expressions in an end-to-end manner based on a data-driven approach.
Experiments demonstrate the efficacy of our model.
In the future, we would like to investigate how such an approach can be applied to more complicated math word problems, like algebra word problems where a problem usually maps to an equation set.
%In the future, we would like to investigate how such a semantic parsing approach can be applied to more complicated math word problems, like algebra word problems where a problem usually maps to an equation set containing multiple variables.
Another interesting direction is to investigate how to incorporate world knowledge into the graph-based approach to boost the performance.

\section*{Acknowledgments}
We would like to thank the three anonymous reviewers for their thoughtful and constructive comments.
This work is supported by Singapore Ministry of Education Academic Research Fund (AcRF) Tier 2 Project MOE2017-T2-1-156.

\bibliography{math}
\bibliographystyle{acl_natbib}

\end{document}